\documentclass[11pt,a4paper]{article}
\usepackage[hyperref]{emnlp2020}
\usepackage{times}
\usepackage{latexsym}

\usepackage{microtype}

\aclfinalcopy 
\def\aclpaperid{858} 

\usepackage{amsmath}
\usepackage{cleveref}

\usepackage{amsfonts}
\usepackage{xcolor}

\usepackage[pdftex]{graphicx}

\usepackage{tabu}
\usepackage{booktabs}
\usepackage{multirow}

\usepackage{subcaption}

\title{Don't Use English Dev: On the Zero-Shot Cross-Lingual Evaluation of Contextual Embeddings}

\author{Phillip Keung \quad Yichao Lu \quad Julian Salazar \quad Vikas Bhardwaj \\
  Amazon \\
  \texttt{\{keung,yichaolu,julsal,vikab\}@amazon.com} \\}

\date{}

\begin{document}
\maketitle

\begin{abstract}
Multilingual contextual embeddings have demonstrated state-of-the-art performance in zero-shot cross-lingual transfer learning, where multilingual BERT is fine-tuned on one source language and evaluated on a different target language. However, published results for mBERT zero-shot accuracy vary as much as 17 points on the MLDoc classification task across four papers. We show that the standard practice of using English dev accuracy for model selection in the zero-shot setting makes it difficult to obtain reproducible results on the MLDoc and XNLI tasks. English dev accuracy is often uncorrelated (or even anti-correlated) with target language accuracy, and zero-shot performance varies greatly at different points in the same fine-tuning run and between different fine-tuning runs. These reproducibility issues are also present for other tasks with different pre-trained embeddings (e.g., MLQA with XLM-R). We recommend providing \emph{oracle} scores alongside zero-shot results: still fine-tune using English data, but choose a checkpoint with the target dev set. Reporting this upper bound makes results more consistent by avoiding arbitrarily bad checkpoints.
\end{abstract}

\section{Introduction}

Zero-shot and zero-resource cross-lingual NLP has seen significant progress in recent years. The discovery of cross-lingual structure in word embedding spaces culminated in the work of \citet{ConneauLRDJ18-word}, which showed that unsupervised word translation via adversarial mappings is competitive with supervised techniques. Concurrent work in machine translation also showed that it is possible to achieve non-trivial BLEU scores without any bitext \citep{ArtetxeLAC18-unsupervised,LampleCDR18-unsupervised}. Self-supervised multilingual contextual embeddings like mBERT \citep{devlin-etal-2019-bert} and XLM \citep{LampleC19-cross} have shown remarkably strong performance on cross-lingual named entity recognition, text classification, dependency parsing, and other tasks (e.g., \citealp{pires-etal-2019-multilingual,keung-etal-2019-adversarial,wu-dredze-2019-beto}).

Much of this recent work has demonstrated that mBERT performs very well on zero-shot tasks, superseding prior techniques as the baseline for zero-shot cross-lingual transfer learning. By \textit{zero-shot}, we mean that no parallel text or labeled data from the target language was used during model training, fine-tuning, or hyperparameter search. In this setting, models are  trained on labeled (usually English) text and tested on target (non-English) text. Standard practice prohibits the use of target language data for model selection; the final model is chosen using the English dev set only.

However, we find that zero-shot mBERT results can vary greatly. We present 4 published baselines for zero-shot cross-lingual document classification on MLDoc \citep{schwenk-li-2018-corpus} in Table \ref{table:published-mbert}: 
\begin{table}[ht]
\begin{minipage}{1.0\linewidth}
	\centering
	\small
	\setlength{\tabcolsep}{4pt}
\begin{tabu}{@{}lccccc@{}}
\toprule
\textbf{MLDoc} & En & Fr & Ja & Ru & Zh \\
\midrule 
\citet{eisenschlos-etal-2019-multifit} & 93.2 & \textbf{83.0} & 64.6 & 71.6 & 66.2 \\
\citet{dong-de-melo-2019-robust} & \textbf{94.2} & 80.0 & \textbf{73.2} & 70.7 & 75.4 \\
\citet{keung-etal-2019-adversarial} & \textbf{94.2} & 73.5 & 72.8 & \textbf{73.7} & 76.0 \\
\citet{wu-dredze-2019-beto} & \textbf{94.2} & 72.6 & 56.5 & \textbf{73.7} & \textbf{76.9} \\
\midrule
$\Delta$ & 1.0 & 10.4 & 16.7 & 3.0 & 10.7 \\
\bottomrule
\end{tabu}
\end{minipage}
\caption{Published mBERT baselines on zero-shot cross-lingual text classification on 4 of 7 target languages of MLDoc. On non-English languages, we see disagreements of up to 17\% (absolute) per column. No paper consistently outperforms the results of another.}
\label{table:published-mbert}
\end{table}

\begin{table*}[ht]

\begin{subtable}{1.0\linewidth}
	\centering
    \footnotesize
\begin{tabu}{@{}lcccccccc@{}}
\toprule
\textbf{MLDoc} & Dev En & De & Es & Fr & It & Ja & Ru & Zh \\
\midrule 
Max & 97.7 & 87.6 & 82.3 & 78.4 & 69.2 & 67.7 & 71.7 & 70.2 \\
Min & 96.8 & 81.7 & 77.7 & 63.4 & 63.2 & 60.6 & 64.5 & 62.6 \\
$\Delta$ & 0.9 & \textbf{5.9} & \textbf{4.6} & \textbf{15.0} & \textbf{6.0} & \textbf{7.1} & \textbf{7.2} & \textbf{7.6} \\
\bottomrule
\end{tabu}
\label{table:variance-mldoc}
\end{subtable}
\newline
\vspace*{0.25 cm}
\newline
\begin{subtable}{1.0\linewidth}
	\centering
    \footnotesize
\begin{tabu}{@{}lccccccccccccccc@{}}
\toprule
\textbf{XNLI} & Dev En & Ar & Bg & De & El & Es & Fr & Hi & Ru & Sw & Th & Tr & Ur & Vi & Zh \\
\midrule 
Max & 82.8 & 66.0 & 69.7 & 71.8 & 67.6 & 75.8 & 74.6 & 61.7 & 69.6 & 50.9 & 55.3 & 61.9 & 60.2 & 71.3 & 71.3 \\
Min & 81.9 & 63.3 & 66.8 & 70.0 & 64.8 & 73.6 & 72.9 & 58.4 & 67.3 & 47.8 & 51.0 & 60.3 & 56.3 & 69.2 & 68.6 \\
$\Delta$ & 0.9 & \textbf{2.7} & \textbf{2.9} & 1.8 & \textbf{2.8} & 2.2 & 1.7 & \textbf{3.3} & 2.3 & 3.1 & \textbf{4.3} & 1.6 & \textbf{3.9} & 2.1 & \textbf{2.7} \\
\bottomrule
\end{tabu}
\label{table:variance-xnli}
\end{subtable}
\caption{Zero-shot accuracies over 10 independent mBERT fine-tuning experiments on MLDoc and XNLI. For each run, we computed the zero-shot accuracies using the checkpoint with the best En dev performance. We show the minimum and maximum accuracy for each task across the 10 experiments. En dev scores are within 0.9\% of each other, but non-En test scores vary by much more depending on the language/corpus. $\Delta \ge$ 2.5\% are \textbf{bolded}.}
\label{table:variance}
\end{table*}

Even though the authors report English accuracies which are basically identical, their target language performances are very different. Given that each experiment starts with the same pre-trained mBERT model and MLDoc dataset, it is clear that these cross-lingual results are \textbf{not reproducible}. For the listed target languages, the highest accuracy is up to 3 points better than the next best and up to 17 points better than the worst. We investigate this reproducibility issue in both MLDoc and XNLI \citep{conneau-etal-2018-xnli}, which is another major dataset for evaluating cross-lingual transfer. The variations in published baselines on these and other datasets are summarized in Sec.~\ref{sec:published}.

In Section \ref{sec:variance}, we show that the final zero-shot accuracies \textbf{between} and \textbf{within} independent mBERT training runs are highly variable. Variations over different random seeds are similar in magnitude to those in Table \ref{table:published-mbert}, with variation due to checkpoint selection using English dev being a significant underlying cause. In Section \ref{sec:directional}, we find that in many cases, \textbf{English (En) dev accuracy is not predictive of target language performance}. In fact, for some languages, En dev performance is actually anti-correlated with target language accuracy. 

Poor correlation between En dev and target test accuracy, combined with high variance between independent runs, means that published zero-shot accuracies are somewhat arbitrary. In addition to zero-shot results, \textbf{we recommend reporting \emph{oracle} results}, where one still fine-tunes using En data, but uses the target dev set for checkpoint selection.
\section{Experimental setup}
\label{sec:setup}

We use cased mBERT (110M parameters) for all of our experiments. We illustrate the reproducibility issues in zero-shot cross-lingual transfer through the document classification task in MLDoc\footnote{\href{https://github.com/facebookresearch/MLDoc}{https://github.com/facebookresearch/MLDoc}} and the natural language inference task in XNLI.\footnote{\href{https://cims.nyu.edu/~sbowman/xnli/}{https://cims.nyu.edu/\textasciitilde sbowman/xnli/}} For the MLDoc and XNLI experiments, we used their full En training sets (10k and 393k examples, respectively). Unless stated otherwise, we use their En development sets (1k and 2.5k examples) and non-En test sets (4k and 5k examples). When fine-tuning mBERT, we used a constant learning rate of $2\times 10^{-6}$ with a batch size of 32 for MLDoc, and a constant learning rate of $2\times 10^{-5}$ with a batch size of 32 for XNLI. Checkpoints were saved at regular intervals: one checkpoint for every 2\% of the training corpus processed. Models were trained until convergence based on En dev accuracy. We used the mBERT implementation in GluonNLP \citep{guo2020gluonnlp} and an AWS p3.8xlarge instance for our MLDoc and XNLI runs, which completed in ${\sim}$3 and ${\sim}$12 hours respectively on a single NVIDIA V100 GPU.

\begin{figure*}[ht]
\centering
\includegraphics[width=0.85\columnwidth,trim={0.2cm 0.3cm 0.2cm 0.2cm},clip]{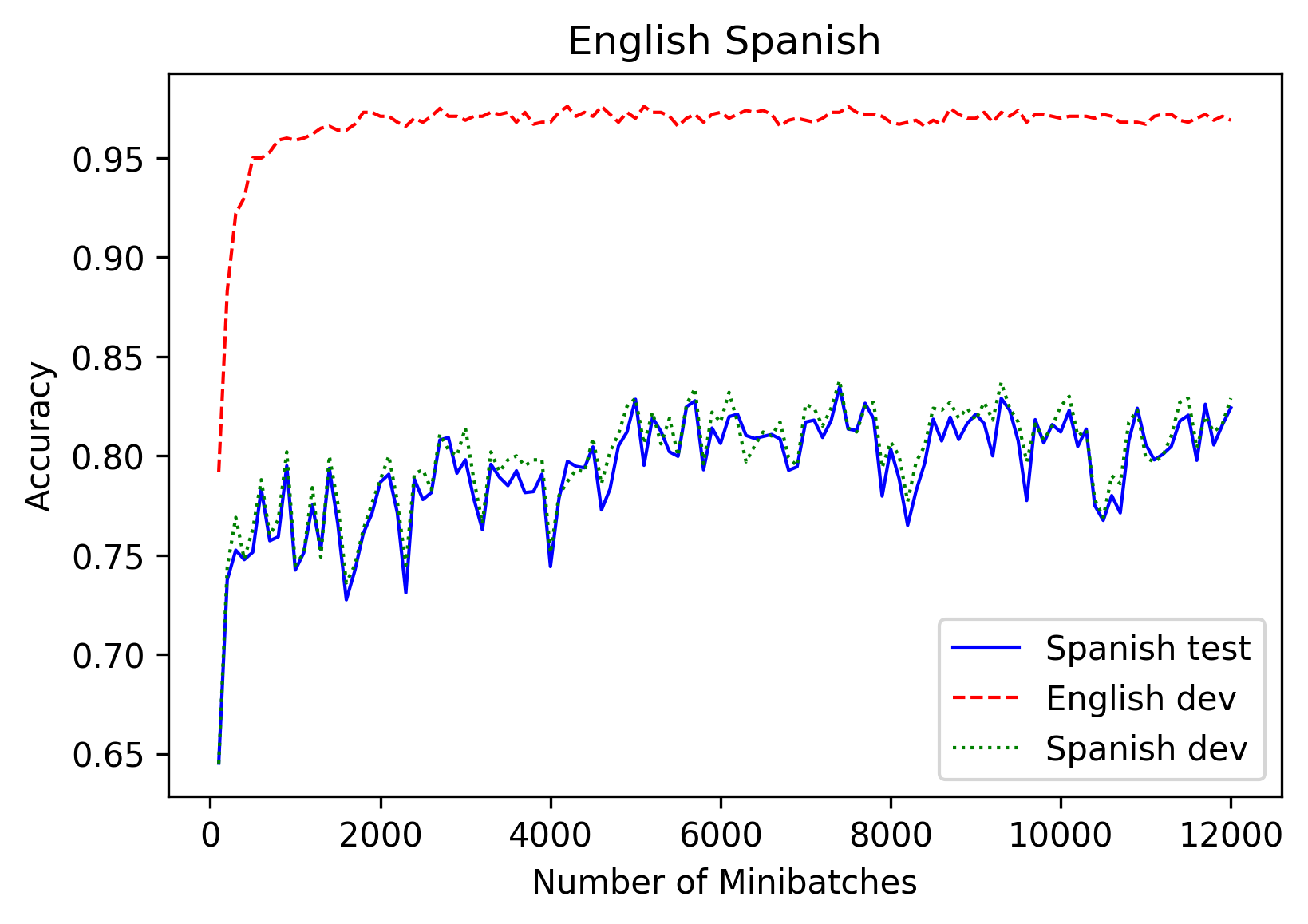}
\hspace{1cm}
\includegraphics[width=0.845\columnwidth,trim={0.2cm 0.3cm 0.2cm 0.2cm},clip]{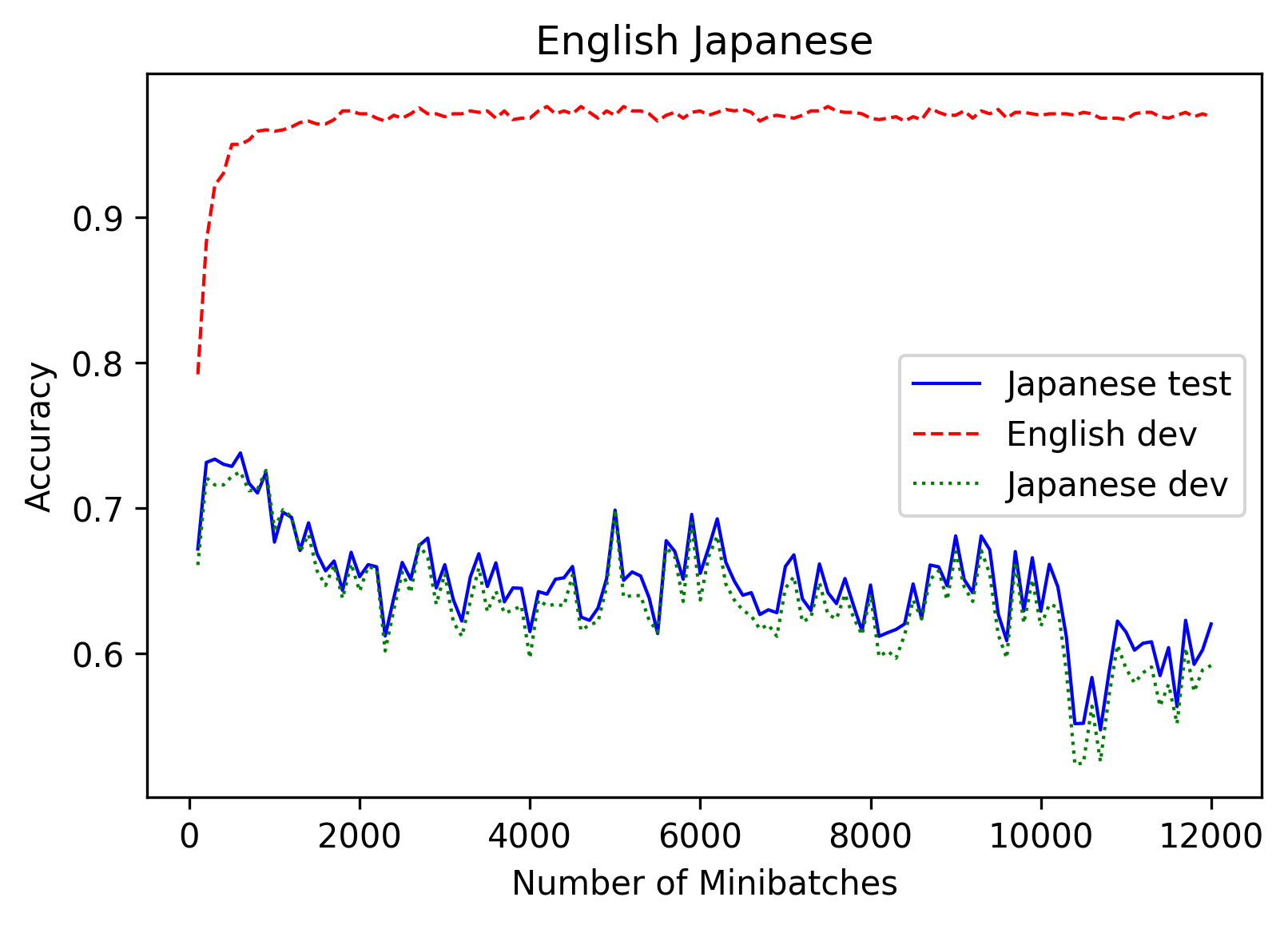}
\caption{En and non-En MLDoc accuracies over a single mBERT fine-tuning run. English and Spanish (59\% directional agreement) accuracy tend to increase together, whereas English and Japanese (42\% directional agreement) accuracy move in opposite directions. Directional agreement below 50\% indicates improvement on the English set at the expense of non-English languages.}
\label{fig:forgetting}
\end{figure*}
\begin{table*}[ht]
\centering
\footnotesize
\begin{tabu}{@{}llcccccccc@{}}
\toprule
& Dev lang. & En & De & Es & Fr & It & Ja & Ru & Zh \\
\cmidrule{2-10}
\textbf{MLDoc} & English & 0.84 & 0.57 & 0.59 & 0.57 & 0.56 & \textbf{0.42} & 0.55 & 0.50 \\
& Target & -- & 0.97 & 0.92 & 0.97 & 0.95 & 0.98 & 0.90 & 0.95 \\
\bottomrule
\end{tabu}

\vspace*{0.25 cm}
\setlength{\tabcolsep}{4.8pt}
\begin{tabu}{@{}llccccccccccccccc@{}}
\toprule
& Dev lang. & En & Ar & Bg & De & El & Es & Fr & Hi & Ru & Sw & Th & Tr & Ur & Vi & Zh \\
\cmidrule{2-17}
\textbf{XNLI} & English & 0.90 & 0.55 & 0.61 & 0.69 & 0.60 & 0.76 & 0.88 & \textbf{0.47} & 0.66 & \textbf{0.44} & 0.56 & \textbf{0.35} & 0.60 & \textbf{0.42} & 0.70 \\
& Target & -- & 0.77 & 0.88 & 0.82 & 0.89 & 0.78 & 0.87 & 0.90 & 0.85 & 0.87 & 0.93 & 0.85 & 0.93 & 0.93 & 0.93 \\
\bottomrule
\end{tabu}

\caption{Frequency of directional agreement between dev and test accuracy on MLDoc and XNLI (higher is better). We expect dev and test accuracy to generally increase and decrease together between randomly sampled checkpoints, which happens when using the target language dev set, but not when using the English dev set. English dev accuracy can be worse than random chance (50\%) at tracking target test accuracy; these values are \textbf{bolded}.}
\label{table:directional}
\end{table*}

\section{Between-run and within-run variation}
\label{sec:variance}

Running mBERT fine-tuning under different random seeds yields highly variable results, similar to what we observed in Table \ref{table:published-mbert}. Previous work on evaluation over random initializations (e.g., \citealp{melis-2018-eval}) reported relatively small effects on the test metric (e.g., $\pm$1 point on En F1 for NER). However, we observed much larger variations in zero-shot accuracy on MLDoc and XNLI.

First, we observed significant variation \emph{between} independent runs (Table \ref{table:variance}). We ran mBERT fine-tuning with different random seeds, and for each run, selected the checkpoint with the best En dev performance. The best checkpoint from each run gave very different zero-shot results, varying as much as 15.0\% absolute in French (Fr) in MLDoc and 4.3\% in Thai (Th) in XNLI.

Second, we observed significant variation \emph{within} each run, which we illustrate in Figure \ref{fig:forgetting}. En dev accuracy reaches a stable plateau as mBERT fine-tuning proceeds; however, zero-shot Spanish (Es) and Japanese (Ja) accuracies swing by several percentage points as training proceeds.

Indeed, for all of the MLDoc languages and for 7 of the 14 XNLI languages, the variation across 10 runs exceeds 2.5\% (absolute), even though the En dev accuracy only varies within a narrow 0.9\% range. To put this variation in context, we note that the MLDoc and XNLI test sets are relatively large, so a 2.5\% difference in accuracy would be statistically significant (at the 5\% significance level using the usual test of proportions), which means that a paper claiming `state-of-the-art' performance on these zero-shot tasks may be reporting strong results because of the large between-run variation, rather than a genuine improvement due to their proposed technique.

In other words, using the En dev accuracy for choosing the final model for zero-shot transfer leads to inconsistent results across different runs. The En dev accuracy is similar for each of our independent experiments, but the target test accuracy for each experiment fluctuates in a wide band.

\begin{table*}[ht]
\begin{subtable}{1.0\linewidth}

\centering
\footnotesize
\setlength{\tabcolsep}{2.1pt}
\begin{tabu}{@{}lccccccccccccccc@{}}
\toprule
\textbf{XNLI} & En & Ar & Bg & De & El & Es & Fr & Hi & Ru & Sw & Th & Tr & Ur & Vi & Zh \\
\midrule
\citet{xtreme} & 80.8 & 64.3 & 68.0 & 70.0 & 65.3 & 73.5 & 73.4 & 58.9 & 67.8 & 49.7 & 54.1 & 60.9 & 57.2 & 69.3 & 67.8 \\
\citet{NooralazadehBBA20-zero} & 81.4 & 64.6 & 67.8 & 69.7 & 65.7 & 73.9 & 73.5 & 58.6 & 67.9 & 47.6 & 52.5 & 59.0 & 58.7 & 70.1 & 68.9 \\
\citet{wu-dredze-2019-beto}, \citet{xglue} & 82.1 & 64.9 & 68.9 & 71.1 & 66.4 & 74.3 & 73.8 & 60.0 & 69.0 & 50.4 & 55.8 & 61.6 & 58.0 & 69.5 & 69.3 \\
\midrule
mBERT $\Delta$ & 1.3 & 0.6 & 1.1 & 1.4 & 1.1 & 0.8 & 0.4 & 0.4 & 1.2 & 2.8 & 3.3 & 2.6 & 1.5 & 0.8 & 1.5 \\
\midrule
\citet{xtreme} & 88.7 & 77.2 & 83.0 & 82.5 & 80.8 & 83.7 & 82.2 & 75.6 & 79.1 & 71.2 & 77.4 & 78.0 & 71.7 & 79.3 & 78.2 \\
\citet{ConneauEtAl19-unsupervised}* & 89.1 & 79.8 & 84.0 & 83.9 & 82.9 & 85.1 & 84.1 & 76.9 & 81.2 & 73.9 & 78.1 & 79.6 & 73.8 & 80.8 & 80.2 \\
\citet{phang-2020-english} & 89.3 & 79.8 & 82.7 & 83.8 & 81.3 & 84.4 & 83.7 & 77.3 & 79.2 & 72.4 & 77.1 & 78.9 & 72.6 & 80.0 & 79.6 \\
\midrule
XLM-R large $\Delta$ & 0.6 & 2.6 & 1.3 & 1.4 & 2.1 & 1.4 & 1.9 & 1.7 & 2.1 & 2.7 & 1.0 & 1.6 & 2.1 & 1.5 & 2.0 \\
\bottomrule
\end{tabu}
\end{subtable}
\newline
\vspace*{0.25 cm}
\newline
\begin{subtable}{1.0\linewidth}
\centering
\footnotesize
\begin{tabu}{@{}lccccccc@{}}
\toprule
\textbf{MLQA} & En & Ar & De & Es & Hi & Vi & Zh \\
\midrule 
\citet{mlqa} & 77.7 & 45.7 & 57.9 & 64.3 & 43.8 & 57.1 & 57.5 \\
\citet{xtreme} & 80.2 & 52.3 & 59.0 & 67.4 & 50.2 & 61.2 & 59.6 \\
\citet{xglue} & 80.5 & 50.9 & 63.8 & 67.1 & 47.9 & 59.5 & 55.4 \\
\midrule
mBERT $\Delta$ & 2.8 & 6.6 & 5.9 & 3.1 & 6.4 & 4.1 & 4.2 \\
\midrule
\citet{ConneauEtAl19-unsupervised}* & 80.6 / 67.8 & 63.1 / 43.5 & 68.5 / 53.6 & 74.1 / 56.0 & 69.2 / 51.6 & 71.3 / 50.9 & 68.0 / 45.4 \\
\citet{phang-2020-english} & 81.6 / 68.6 & 62.7 / 42.4 & 69.1 / 52.0 & 72.2 / 53.0 & 68.0 / 50.7 & 69.5 / 47.6 & 67.9 / 46.2 \\
\citet{xtreme} & 83.5 / 70.6 & 66.6 / 47.1 & 70.1 / 54.9 & 74.1 / 56.6 & 70.6 / 53.1 & 74.0 / 52.9 & 62.1 / 37.0 \\
\midrule
XLM-R large $\Delta$ & 2.9 / 2.8 & 3.9 / 4.7 & 1.6 / 2.9 & 1.9 / 3.6 & 2.6 / 2.4 & 4.5 / 5.3 & 5.9 / 8.4 \\
\bottomrule
\end{tabu}
\end{subtable}

\caption{Published zero-shot accuracies for mBERT and XLM-R large on XNLI, and published zero-shot F1 scores for mBERT and F1 / EM scores for XLM-R large on MLQA. * \citet{ConneauEtAl19-unsupervised} tune on all dev sets jointly.}
\label{table:published-xnli-mlqa}
\end{table*}

\section{English dev accuracy and its relationship with zero-shot accuracy}
\label{sec:directional}

Experimenters use the En dev set for model selection under the assumption that zero-shot performance improves as En dev performance improves. We show that this assumption is often false.

We compare the abilities of En dev and target dev to predict directional changes in target test accuracy. In Table \ref{table:directional}, we report the frequency of \emph{directional agreement} on MLDoc and XNLI; i.e., how often does En dev accuracy increase/decrease in tandem with target test accuracy?

We randomly sample pairs of checkpoints where the change in target test accuracy for the pair is at least 0.5\%, and compute the proportion for which the En dev accuracy changed in the same direction. Table \ref{table:directional} shows that for MLDoc, En dev is not much better than a coin flip ($\sim$50\%) at predicting the direction of the change in target test accuracy, while target dev tracks target test accuracy 90+\% of the time. For XNLI, En dev sometimes approaches target dev in predictive power (i.e., Es and Fr), but can fall short for other languages. In general, we see higher directional agreement in XNLI than MLDoc, which we attribute to XNLI's target test sets being professional translations of En test.

Remarkably, for some languages (i.e., Ja in MLDoc and Hi, Sw, Tr, and Vi in XNLI), the frequency of directional agreement is less than 50\%, which means that, more often than not, when En dev accuracy increases, target test accuracy for these languages decreases; we discuss this in Section \ref{sec:forgetting}. Since En dev accuracy does not reliably move in the same direction as target test accuracy, it is an inadequate metric for tracking zero-shot cross-lingual transfer performance.

\section{Catastrophic forgetting}
\label{sec:forgetting}

The strange phenomenon in Table \ref{table:directional} (where the probability of directional agreement is sometimes less than 50\%) occurs even on XNLI, where the dev and test sets are translated from English and therefore have the same semantic content. We believe this phenomenon is a form of catastrophic forgetting \citep{kirkpatrick2017forget}, where mBERT loses some cross-lingual knowledge from pre-training because it is fine-tuned on En data only.

In Figure \ref{fig:forgetting}, we plotted the En dev accuracy and the target test accuracy over time, for the language with the highest directional agreement (Es, 0.59) and for the language with the lowest directional agreement (Ja, 0.42) for MLDoc (see Table \ref{table:directional}). From the figure, Es test accuracy does increase with En dev accuracy, while Ja test accuracy decreases as En dev accuracy increases. The same pattern holds with XNLI for Tr and En (not shown), where Turkish accuracy decreases somewhat as fine-tuning with English training data continues.

We conclude that En dev accuracy cannot detect when mBERT is improving on the En training data at the expense of non-En languages, and should not (solely) be used to assess zero-shot performance.

\begin{table*}[ht]
\begin{subtable}{1.0\linewidth}
	\centering
    \footnotesize
\begin{tabu}{@{}lccccccc@{}}
\toprule
\textbf{MLDoc} & De & Es & Fr & It & Ja & Ru & Zh \\
\midrule
Target dev (oracle) & \textbf{89.7} & \textbf{84.4} & \textbf{84.4} & \textbf{73.1} & \textbf{75.5} & \textbf{77.1} & \textbf{81.1} \\
Best En dev & 87.6 & 82.3 & 78.4 & 69.2 & 67.7 & 71.7 & 70.2 \\
Best published & 82.4 & 79.5 & 83.0 & 68.9 & 73.2 & 73.7 & 76.9 \\
\bottomrule
\end{tabu}
\label{table:oracle-mldoc}
\end{subtable}
\newline
\vspace*{0.25 cm}
\newline
\begin{subtable}{1.0\linewidth}
	\centering
    \footnotesize
    \setlength{\tabcolsep}{5.5pt}
\begin{tabu}{@{}lcccccccccccccc@{}}
\toprule
\textbf{XNLI} & Ar & Bg & De & El & Es & Fr & Hi & Ru & Sw & Th & Tr & Ur & Vi & Zh \\
\midrule 
Target dev (oracle) & \textbf{66.5} & \textbf{70.0} & \textbf{72.0} & \textbf{67.8} & \textbf{75.9} & \textbf{74.6} & \textbf{63.2} & \textbf{70.7} & \textbf{52.9} & \textbf{57.3} & \textbf{63.0} & \textbf{60.5} & \textbf{71.4} & \textbf{71.3} \\
Best En dev & 66.0 & 69.7 & 71.8 & 67.6 & 75.8 & 74.6 & 61.7 & 69.6 & 50.9 & 55.3 & 61.9 & 60.2 & 71.3 & 71.3 \\
Best published & 64.9 & 68.9 & 71.1 & 66.4 & 74.3 & 73.8 & 60.0 & 69.0 & 50.4 & 55.8 & 61.6 & 58.7 & 70.1 & 69.3 \\

\bottomrule
\end{tabu}
\label{table:oracle-xnli}
\end{subtable}

\caption{Oracle zero-shot accuracies with mBERT across 10 independent runs, using target dev to select the best checkpoint for each language. This provides an upper bound on the achievable zero-shot accuracy. Published results are derived from sources in Tables \ref{table:published-mbert}, \ref{table:published-xnli-mlqa}, and \ref{table:published-mldoc}. Best En dev results are from Table \ref{table:variance}.}
\label{table:oracle-dev}
\end{table*}

\section{Published results}
\label{sec:published}

At the time of writing, over 70 papers have been published on the MLDoc and XNLI corpora. Several new datasets (e.g., \citealp{xqa, xquad, pawsx}) for zero-shot cross-lingual evaluation have been released and aggregated into benchmark suites like XGLUE \citep{xglue} and XTREME \citep{xtreme}. There are also more recent multilingual contextual embeddings (e.g., XLM-R large; \citealp{ConneauEtAl19-unsupervised}) which is trained on significantly more data (2.5 terabytes). Notably, \citet{ConneauEtAl19-unsupervised} uses the target dev sets \textit{jointly} for checkpoint selection. 

As with Table~\ref{table:published-mbert} for MLDoc, we search for signs of the variability found in our experimental setup across the zero-shot cross-lingual literature.  In XNLI (Table~\ref{table:published-xnli-mlqa}) we see less drastic variations in $\Delta$. However, for variations in En of 0.6 (XLM-R) and 1.3 (mBERT) points, we see variations of $\ge$2.5 points in linguistically distant languages (Ar, Sw, Th, Tr), languages which agreed with En dev $\le$56\% during finetuning (Table~\ref{table:directional}). We also include MLQA (\citealp{mlqa}), where a variation of 2.8 in En mBERT results gave even higher variations ($\ge$4.0 points) in all languages except Es. A similar effect occurs with XLM-R and the more distant languages (Ar, Vi, Zh). In particularly, Chinese (Zh) degrades by 5.9 points as En improves, which may be evidence of directional disagreement in MLQA between En dev and Zh test.

In Table~\ref{table:published-ner}, results on CoNLL 2002/2003 (\citealp{sang2003introduction}) show slightly increased variation relative to En. We also include the remaining MLDoc languages for completeness.\footnote{The English result from \citet{eisenschlos-etal-2019-multifit} is from \href{https://github.com/n-waves/multifit/commit/b7e3a5e7a46c774ed00cbc94bb159bec86bf8fa6}{b7e3a5 of https://github.com/n-waves/multifit}.}

\begin{table}[ht]
\begin{subtable}{1.0\linewidth}
\centering
\footnotesize
\begin{tabu}{@{}lcccc@{}}
\toprule
\textbf{CoNLL 2002/2003} & En & De & Es & Nl \\
\midrule 
\citet{xglue} & 90.6 & 69.2 & 75.4 & 77.9 \\
\citet{pires-etal-2019-multilingual} & 90.7 & 69.7 & 73.6 & 77.4 \\
\citet{keung-etal-2019-adversarial} & 91.1 & 68.6 & 75.0 & 77.5 \\
\citet{bari-2020-multimix} & 91.1 & 71.0 & 74.8 & 79.6 \\
\citet{wu-dredze-2019-beto} & 92.0 & 69.6 & 75.0 & 77.6 \\
\midrule
mBERT $\Delta$ & 1.4 & 2.4 & 1.8 & 2.2 \\
\bottomrule
\end{tabu}
\end{subtable}
\newline
\vspace*{0.25 cm}
\newline
\begin{subtable}{1.0\linewidth}
\centering
\footnotesize
\begin{tabu}{@{}lcccc@{}}
\toprule
\textbf{MLDoc (cont.)} & En & De & Es & It \\
\midrule 
\citet{eisenschlos-etal-2019-multifit} & 93.2 & 82.4 & 75.0 & 68.3 \\
\citet{dong-de-melo-2019-robust} & 94.2 & 78.9 & 79.5 & 68.7 \\
\citet{keung-etal-2019-adversarial} & 94.2 & 79.8 & 72.1 & 63.7 \\
\citet{wu-dredze-2019-beto} & 94.2 & 80.2 & 72.6 & 68.9 \\
\midrule
mBERT $\Delta$ & 1.0 & 3.5 & 7.4 & 5.2 \\
\bottomrule
\end{tabu}
\end{subtable}
\caption{Published zero-shot F1 scores for mBERT on cross-lingual NER (CoNLL 2002/2003), and published zero-shot accuracies for mBERT on MLDoc for languages not included in Table 1.}
\label{table:published-ner}
\label{table:published-mldoc}
\end{table}
\section{Recommendations and discussion}
\label{sec:discussion}

Using a poor metric like En dev accuracy to select a model checkpoint is similar to picking a checkpoint at random. This would not be a major issue if the variance between different training runs were low; the test performance would, in that case, be consistently mediocre. The problem arises when the variability is high, which we have seen experimentally (Table \ref{table:variance}) and in the wild (Table \ref{table:published-mbert}).

We showed that independent experiments can report very different results, which prevents us from making meaningful comparisons between different baselines and methods. Currently, it is standard practice to use the En dev accuracy for checkpoint selection in the zero-shot cross-lingual setting. However, we showed that using En dev accuracy for checkpoint selection leads to somewhat arbitrary zero-shot results.

Therefore, we propose reporting \emph{oracle} accuracies, where one still fine-tunes using English data, but selects a checkpoint using target dev. This represents the \textbf{maximum achievable zero-shot accuracy}. Note that we do not use target dev for hyperparameter tuning; we are using target dev to avoid selecting bad checkpoints within each fine-tuning experiment. Table \ref{table:oracle-dev} shows our oracle results on MLDoc and XNLI. Reporting this upper bound makes results more consistent by avoiding arbitrarily bad checkpoints. To avoid unexpected variation in future cross-lingual publications, we recommend that authors report oracle accuracies alongside their zero-shot results.

\bibliography{paper_all}
\bibliographystyle{acl_natbib}

\end{document}